\documentclass[11pt,a4paper]{article}
\usepackage[hyperref]{acl2020}
\usepackage{amsmath}
\usepackage{times}
\usepackage{latexsym}
\usepackage{multirow}
\usepackage{amssymb}
\usepackage{graphicx} 

\usepackage{subcaption}
\usepackage{xcolor}
\usepackage{microtype}

\aclfinalcopy % Uncomment this line for the final submission
\def\aclpaperid{2131} %  Enter the acl Paper ID here

\title{Improving Segmentation for Technical Support Problems}

\author{Kushal Chauhan \thanks{~~Work done at IBM Research during a summer internship} \\
  ABV-IIITM, Gwalior \\
  \texttt{kushalchauhan98@gmail.com} \\\And
  Abhirut Gupta \thanks{~~Now at Google}\\
  IBM Research AI \\
  \texttt{abhirut.91@gmail.com} \\}

\date{}

\begin{document}
\maketitle
\begin{abstract}
  Technical support problems are often long and complex. They typically contain user descriptions of the problem, the setup, and steps for attempted resolution. Often they also contain various non-natural language text elements like outputs of commands, snippets of code, error messages or stack traces. These elements contain potentially crucial information for problem resolution. However, they cannot be correctly parsed by tools designed for natural language. In this paper, we address the problem of segmentation for technical support questions. We formulate the problem as a sequence labelling task, and study the performance of state of the art approaches. We compare this against an intuitive contextual sentence-level classification baseline, and a state of the art supervised text-segmentation approach. We also introduce a novel component of combining contextual embeddings from multiple language models pre-trained on different data sources, which achieves a marked improvement over using embeddings from a single pre-trained language model. Finally, we also demonstrate the usefulness of such segmentation with improvements on the downstream task of answer retrieval.
\end{abstract}

\section{Introduction}
\begin{figure}[t]
    \centering
    \includegraphics[width=0.45\textwidth]{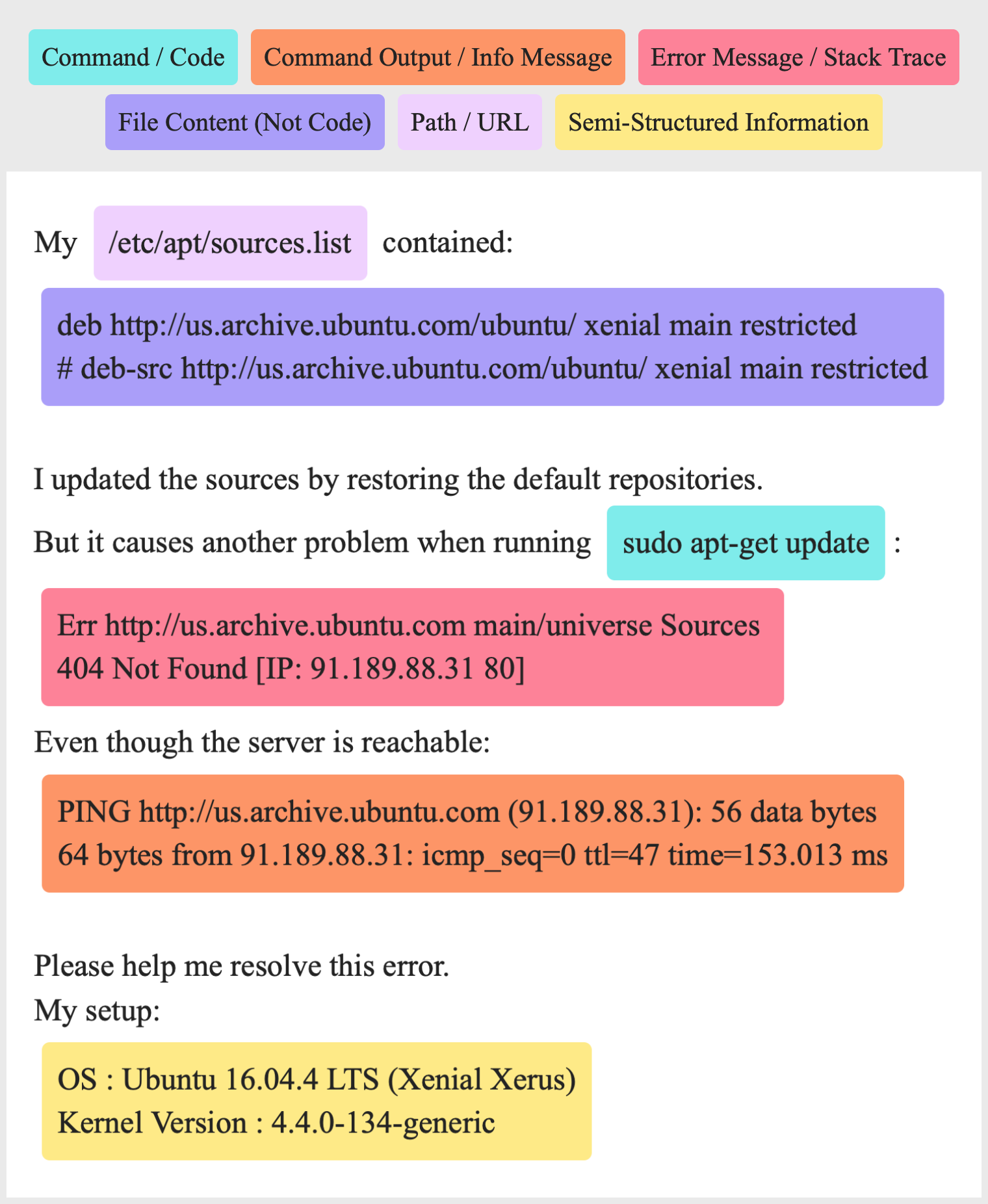}
    \caption{Various non-natural language segments labelled from a problem on AskUbuntu}
    \label{fig:SegmentLabels}
\end{figure}

% \todoag{Introduce the ``technical support" setup, cite different work in the domain, cite gap for such syntactic segmentation.}
Problems, reported by users of software or hardware products - called \textit{tickets} or \textit{cases}, are often long and complex. Along with a description of the problem, users often report the setup, steps they have tried at mitigating the problem, and explicit requests. These problems also contain various non-natural language elements like snippets of code or commands tried, outputs of commands or software tools, error messages or stack traces, contents of log files or configuration files, and lists of key-value pairs. Figure~\ref{fig:SegmentLabels} shows a sample support problem from AskUbuntu~\footnote{\href{https://askubuntu.com/}{https://askubuntu.com/}}, where all such segments are labeled.

% While they're important sources of information to the human reader, they are difficult to handle for systems designed to automatically answer them. As noted in ~\citealt{gupta-etal-2018-semantic}, these segments lead to parsing mistakes, and errors in 
While these segments are important sources of information for the human reader, they are difficult to handle for systems built to automatically answer support problems. As noted in~\citet{gupta-etal-2018-semantic}, the non-natural language segments lead to parsing mistakes, and errors in the understanding of support problems. Correctly identifying these segments can also augment problem understanding. For instance, a retrieval engine with error messages and their solutions indexed in distinct fields would return better results with a fielded query containing just the error message from the ticket. Specialized tools for log analysis~\cite{he2016experience} could also be run specifically on the identified log segment of problems.

% \todokc{Give example ticket with labels (from poster) and explain the labels}
% \begin{figure*}[htp]
% \centering
% \begin{minipage}[b]{.47\textwidth}
% \includegraphics[scale=0.17]{images/code_error}
% \end{minipage}\qquad
% \begin{minipage}[b]{.47\textwidth}
% \includegraphics[scale=0.17]{images/code_config}
% \end{minipage}
% \caption{Sample problems from Ask Ubuntu with $\langle code \rangle$ tag used to present (left) an error message, and (right) contents of a configuration file}
% \label{fig:codeWrong}
% \end{figure*}

\begin{figure*}[h]

\begin{subfigure}[b]{\textwidth}
    \centering
   \includegraphics[scale=0.25]{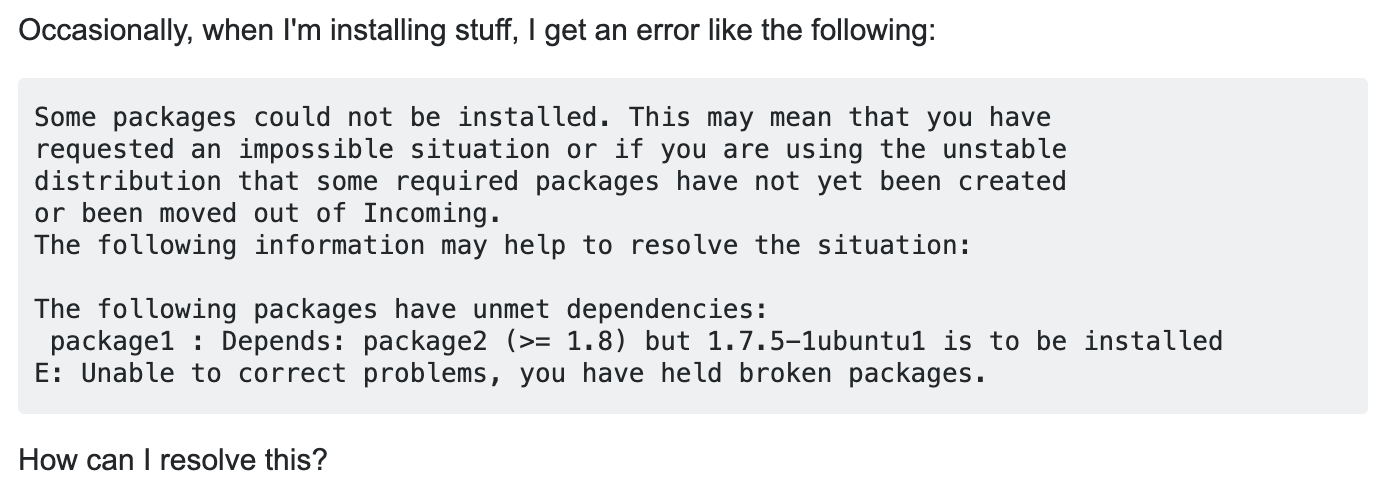}
   \caption{}

\end{subfigure}

\begin{subfigure}[b]{\textwidth}
\centering
   \includegraphics[scale=0.25]{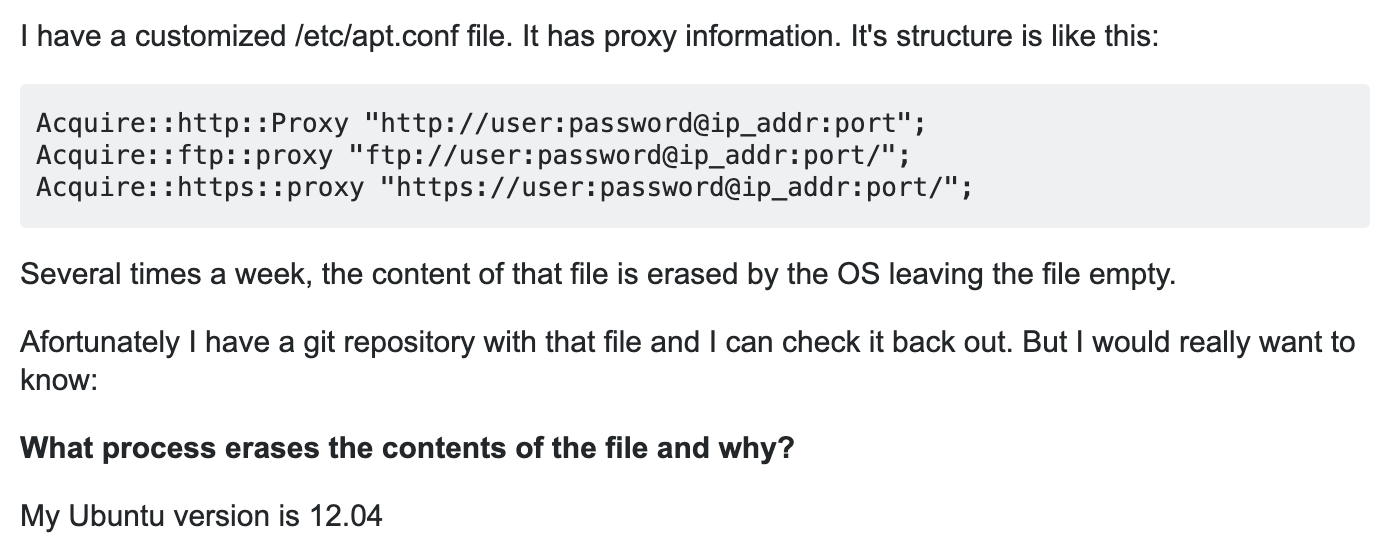}
   \caption{}

\end{subfigure}

\caption{Sample problems from Ask Ubuntu with $\langle code \rangle$ tag used to present (a) an error message, and (b) contents of a configuration file}
\label{fig:codeWrong}
\end{figure*}

In this paper, we aim to address the problem of identifying and extracting these non-natural language segments from support tickets. In particular, we choose to focus on the following six segment labels which appear often in support tickets (also shown in Figure~\ref{fig:SegmentLabels}):
\begin{itemize}
    \item \textbf{Command / Code}: Includes terminal commands and programming code snippets
    \item \textbf{Command Output / Info Message}: Includes outputs of successful command/code executions
    \item \textbf{Error Message / Stack Trace}: Includes error traces resulting from unsuccessful command/code executions
    \item \textbf{File Content (Not Code)}: Includes contents of log files, configuration files, etc. which do not contain programming source code
    \item \textbf{Path / URL}: Includes file paths or webpage URLs
    \item \textbf{Semi-Structured Information}: Includes text which is structured in the form of key-value pairs, lists, etc., often used to convey system configurations or lists of components
\end{itemize}

We formulate the problem as a sequence labelling task, with word-level tags used to encode segments. To leverage the rich literature of supervised approaches in this framework, we also create a dataset with segments tagged for questions from AskUbuntu~\footnote{Data available at \href{https://github.com/kushalchauhan98/ticket-segmentation}{https://github.com/kushalchauhan98/ticket-segmentation}}. 
%We would like to point out that while posts on the website support the $\langle code \rangle$ html tag, it is not granular enough for our downstream tasks. These tags are also often abused to present snippets of command outputs/error messages/file paths etc. Figure~\ref{fig:codeWrong} shows examples of such questions.

% \begin{figure*}[htp]
%   \centering
%   \subfigure[random caption 1]{\includegraphics[scale=0.38]{images/code_error}}\quad
%   \subfigure[random caption 2]{\includegraphics[scale=0.38]{images/code_config}}
% \end{figure*}

% In total, we fetch tags for over 1,300 questions. However, given the verbosity of these questions, it corresponds to over 1.18 million word-level tokens. To the best of our knowledge, this is the only such dataset with granular segments.

% We explore various components of popular RNN based architectures for sequence labelling, on this task. In particular, we study the effect of different pre-trained word representations, learned character-based embeddings for words, and the attention layer. Pre-trained contextual embeddings are also explored, and we find they improve performance on the task.

% We also introduce the novel idea of combining pre-trained contextual embeddings from language models trained on different data sources (English text, code repositories, configuration files, etc.). We show this method achieves a significant improvement over using embeddings from a single pre-trained language model.

Our contributions are as follows -
\begin{enumerate}
    \item We introduce a novel task towards understanding technical support problems, which has implications on a variety of downstream applications. We also release a tagged dataset of problems for the task.%~\footnote{Data available at \href{https://github.com/kushalchauhan98/ticket-segmentation}{https://github.com/kushalchauhan98/ticket-segmentation}}.
    \item We benchmark the performance of state of the art sequence labelling models on the task, studying their performance and limitations. This hopefully provides direction for future research.
    \item Given the relatively small size of tagged data, we also explore pre-training based approaches. Our model leverages activations from multiple language models pre-trained on different data sources, and we show how they can be used to improve performance on the task.
    
\end{enumerate}

\section{Related Work}
%1. \todoag{Work on tech support problems - cite the Coling paper here - talk about how that's "semantic segmentation" but this is syntacting.}
%\todokc{Also cite CQA segmentation of questions and their related context sentences - slightly un-related task. Also cite paper that does code-comment segmentation using the <code> tag for data - people don't use such tags often, and insufficient granularity for downstream tasks.}

Understanding technical support problems is a particularly difficult task, owing to the long text of problems. In~\citet{gupta-etal-2018-semantic}, the authors propose that understanding can be approached by extracting attributes of the ticket that correspond to the description of the problem (symptom), steps taken for mitigation (attempt), and explicit requests (intent). They also propose a dependency parser-based approach for extracting these attributes. However, while this approach pays attention to the semantics of the problem, the syntactical idiosyncrasies are ignored. 
%In fact, one of the drawbacks identified in the paper was that these non-natural language segments lead to mistakes in the dependency parse, leading to poor extraction of these attributes.

The idea of segmenting of questions for improvements on downstream tasks is not new. In \citet{Wang2010}, the authors propose an unsupervised graph-based approach for segmenting questions from Community Question Answering (cQA) websites into sub-questions and their related context sentences. The authors demonstrate improvements in question retrieval by using these segments for more granular similarity matching. \citet{Chrupala2013} uses representations from a character-level language model for segmenting code spans in Stack Overflow posts. The author uses $\langle code \rangle$ tags in HTML sources of posts for supervised training of a character level sequence labelling model. However, the $\langle code \rangle$ tags in the posts usually include all forms of non-natural language text like code snippets, command outputs, error messages or stack traces, and file paths (See Fig~\ref{fig:codeWrong}). The resulting level of granularity is thus insufficient for effective application in downstream tasks such as automated problem resolution. The task of text-segmentation in itself has been well studied in the literature, with popular unsupervised approaches like TextTiling~\cite{Hearst1997TextTilingST} and C99~\cite{choi-2000-advances}. While, the problem of ticket segmentation, as defined by us, involves both segmenting and identifying segment types, we compare the performance of a more recent supervised segmentation approach~\cite{DBLP:journals/corr/abs-1803-09337} against our proposed model.

%2. \todokc{Work on sequence labelling for text segmentation.}
Significant amount of work has been done on using sequence labelling approaches for text segmentation tasks \cite{Huang2015,Chiu2016,Lample2016,Ma2016,Rei2016,Peters2017}. 
%\citet{Huang2015} proposes a biLSTM-CRF based model for sequence labelling and achieves state of the art performance on chunking, NER, and POS tagging tasks. In \citet{Chiu2016}, \citet{Lample2016}, \citet{Ma2016}, the authors use character-level word representations in combination with pre-trained word embeddings as an input to a biLSTM-CRF based architecture for sequence labelling. \citet{Rei2016} demonstrates that using attention over word and character level features instead of naive concatenation can improve performance. 
%To compensate for the relatively small size of most sequence labelling corpora, transfer learning, and multi-task learning techniques have received much attention. \citet{Peters2017} proposes the use of word representations from a pre-trained bidirectional language model as an input to assist in the supervised training in chunking and NER tasks. 
In \citet{Wang2018} the authors use ELMo embeddings and a biLSTM-CRF based architecture with self-attention for the task of neural discourse segmentation. We adopt a similar architecture, and explore the effect of using pre-trained contextual embeddings on our task. Given the fact that different segments in technical support problems have very different vocabularies, we also explore leveraging pre-trained Language Models on a variety of different datasets.

%\citet{Akbik2018} uses word representations from a pre-trained character-level language model for sequence labelling and achieves a new state of the art for the NER task. \citet{Rei2017} introduces language modelling as a secondary training objective in sequence labelling tasks to learn more general patterns that can be reused in the primary sequence labelling objective. Realising that discrepancy between the language modelling and sequence labelling tasks may hurt performance, \citet{Liu2018} uses separate highway layers for each of the language modelling and sequence labelling training objectives on top of character level inputs. %Given the fact that different segments in technical support problems have very different vocabularies, we explore leveraging pre-trained Language Models on a variety of different datasets.

\section{Data}

% \begin{table*}[t]
% \centering
% \begin{tabular}{|l|l|l|l|l|l|l|l|l|l|}
% \hline
%      & \textbf{\#Questions} & \textbf{Avg. \#Words} & \multicolumn{7}{|l|}{\textbf{Avg. \#Spans}}\\
%      & & & \textbf{Total} & \textbf{CC}  & \textbf{CO}  & \textbf{ES}  & \textbf{FC}  & \textbf{SS} & \textbf{PU} \\
% \hline
%     Dataset & 1317 & 897.37 & 4.86 & 2.13 & 1.20 & 0.62 & 0.30 & 0.14 & 0.46\\
%     \hspace{10pt}Train & 1053& 899.33 & 4.91 & 2.14 & 1.20 & 0.63 & 0.30 & 0.14 & 0.49\\
%     \hspace{10pt}Val & 131& 783.43 & 4.67 & 2.17 & 1.04 & 0.66 & 0.26 & 0.19& 0.36\\
%     \hspace{10pt}Test & 133& 994.10 & 4.64 & 2.08 & 1.36 & 0.47 & 0.35 & 0.09 & 0.28\\
% \hline
% \end{tabular}
% \caption{Statistics of the tagged dataset for segmentation with average number of words and spans per question. The last 6 columns contain average number of spans for each tag type - \textbf{CC}: Command/Code, \textbf{CO}: Command Output, \textbf{ES}: Error Message/Stack Trace, \textbf{FC}: File Content, \textbf{SS}: Semi-structured Information, \textbf{PU}: Path/URL}
% \label{dataset}
% \end{table*}

\begin{table*}[t]
\centering
\begin{tabular}{|l|l|l|l|l|l|l|l|l|l|}
\hline
     & \multirow{2}{*}{\textbf{\#Questions}}  & \multirow{2}{*}{\textbf{Avg. \#Words}} & \multicolumn{7}{|c|}{\textbf{Avg. \#Spans}}\\
     \cline{4-10}
     & & & \textbf{Total} & \textbf{CC}  & \textbf{CO}  & \textbf{ES}  & \textbf{FC}  & \textbf{SS} & \textbf{PU} \\
\hline
    Dataset & 1317 & 897.37 & 4.86 & 2.13 & 1.20 & 0.62 & 0.30 & 0.14 & 0.46\\
    \hspace{10pt}Train & 1053& 899.33 & 4.91 & 2.14 & 1.20 & 0.63 & 0.30 & 0.14 & 0.49\\
    \hspace{10pt}Val & 131& 783.43 & 4.67 & 2.17 & 1.04 & 0.66 & 0.26 & 0.19& 0.36\\
    \hspace{10pt}Test & 133& 994.10 & 4.64 & 2.08 & 1.36 & 0.47 & 0.35 & 0.09 & 0.28\\
\hline
\end{tabular}
\caption{Statistics of the tagged dataset for segmentation with average number of words and spans per question. The last 6 columns contain average number of spans for each tag type - \textbf{CC}: Command/Code, \textbf{CO}: Command Output, \textbf{ES}: Error Message/Stack Trace, \textbf{FC}: File Content, \textbf{SS}: Semi-structured Information, \textbf{PU}: Path/URL}
\label{dataset}
\end{table*}

% \begin{table}[t]
% \centering
% \begin{tabular}{|p{0.35\textwidth}|l|}
% \hline
%      \textbf{Label} & \textbf{Length} \\
% \hline
%     Command / Code & 12.69 \\
%     Command Output / Info Message & 202.71 \\
%     Error Message / Stack Trace & 251.04\\
%     File Content & 258.68\\
%     Path / URL & 163.36\\
%     Semi-Structured Information & 4.46\\
% \hline
% \end{tabular}
% \caption{Average length of each span (in words)}
% \label{spanlengths}
% \end{table}

\begin{figure}[t]
    \centering
    \includegraphics[width=0.45\textwidth]{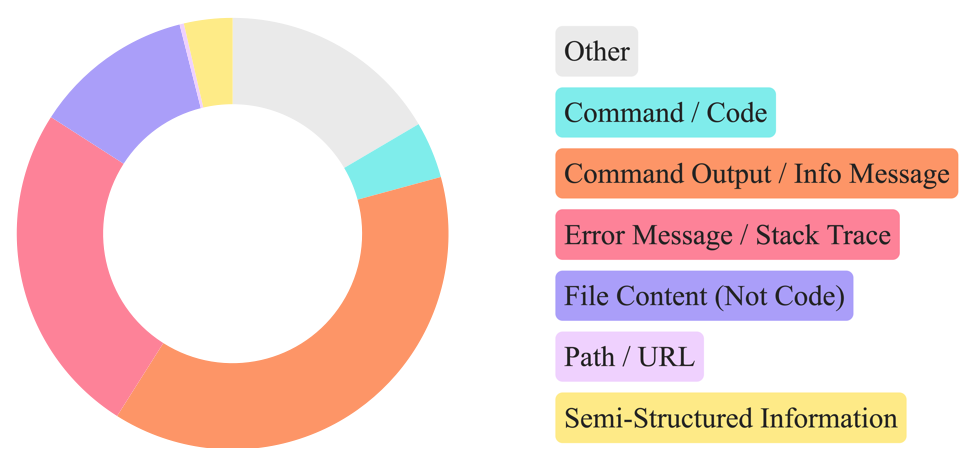}
    \caption{Relative frequencies of each tag in the dataset.}
    \label{fig:pie}
\end{figure}

\begin{figure}[t]
    \centering
    \includegraphics[width=0.48\textwidth]{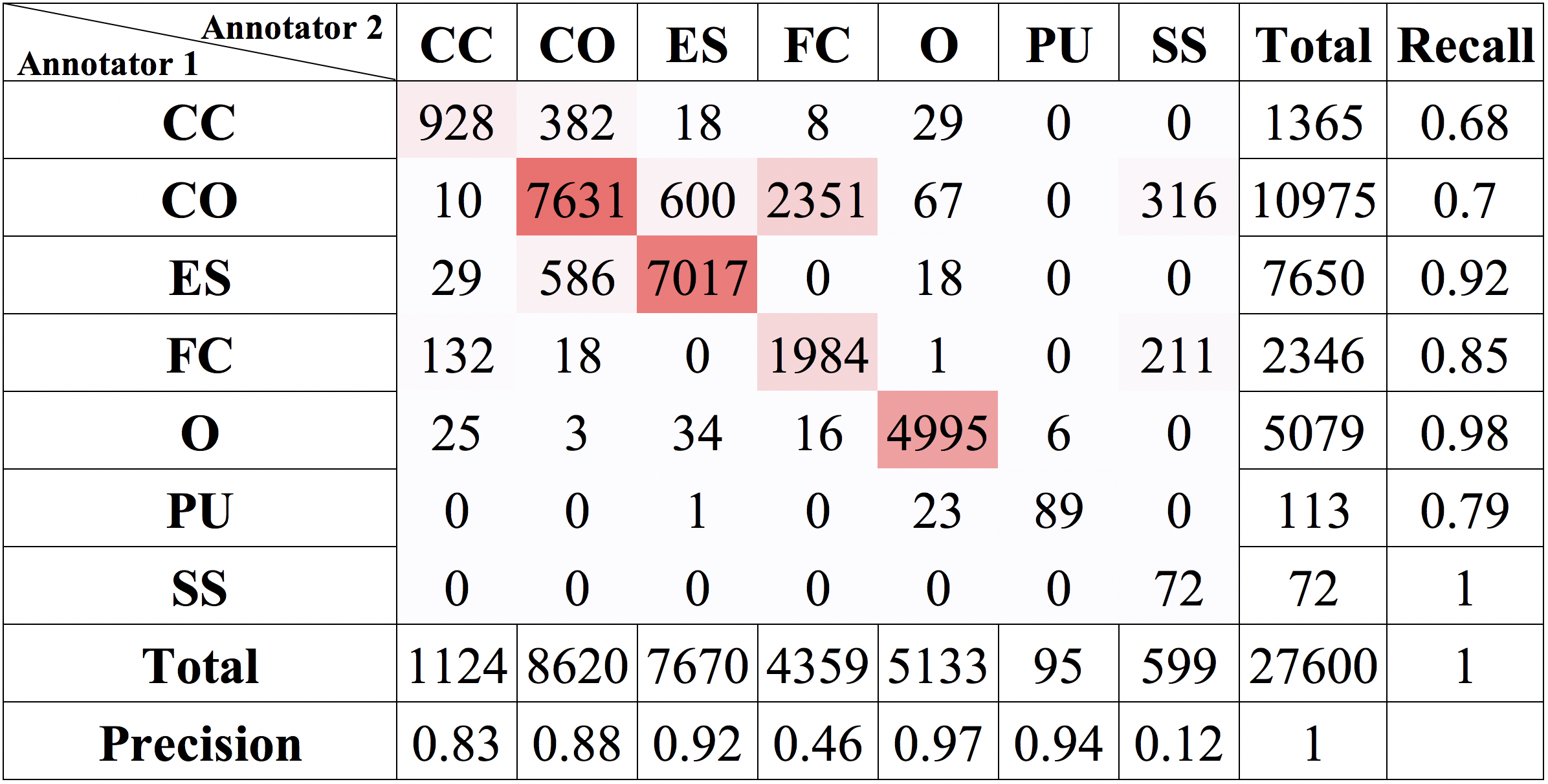}
    \caption{Confusion Matrix to show the word-level  agreement between annotations of 2 annotators on 50 questions. The relatively large off-diagonal values represent the inherent difficulty in the task. Abbreviations for tags - \textbf{CC}: Command/Code, \textbf{CO}: Command Output, \textbf{ES}: Error Message/Stack Trace, \textbf{FC}: File Content, \textbf{SS}: Semi-structured Information, \textbf{PU}: Path/URL}
    \label{fig:confusion_annotator}
\end{figure}

Our dataset is derived from English questions on Ask Ubuntu. Questions posted on the website are similar to proprietary tech support tickets (in terms of question length, number of keywords/noun phrases, etc). We would like to point out that while posts on the website support the $\langle code \rangle$ HTML tag, it is not granular enough for our downstream tasks. These tags are also often abused to present snippets of command outputs/error messages/file paths etc. Figure~\ref{fig:codeWrong} shows examples of such questions. We also do not use other metadata available (like turn-based information) with the data dump because these are not available with proprietary tickets.

%For collecting and annotating data, we concentrate on questions in English from the websites Ask Ubuntu and Super User, since these questions have similar properties (length of questions, number of keywords/noun phrases etc.) to proprietary Tech Support tickets. 
% As a first pre-processing step, we disregard the markup information offered by the $\langle code \rangle$ HTML tag from posts on these websites, since they often contain command outputs, error messages, file paths, apart from code snippets. Figure~\ref{fig:codeWrong} shows examples of such questions. These tags are thus not granular for our downstream task. We are left with a complete text dump of the questions containing all the segments of interest, without any markup. Next, we shortlist questions that are likely to contain non-natural language segments. This is done to avoid wasting annotator time. We select all tickets that have at least 20\% tokens from outside an English dictionary. 

% We now focus on eliciting tags for these questions.
Tagging is performed at the word level, and we use the BIO tagging scheme. We have a pair of Begin and Inside tags for each of the 6 non-natural language segments, and natural language segments are labelled O, totalling to 13 tags. We use the Doccano tool~\footnote{\href{https://github.com/chakki-works/doccano}{https://github.com/chakki-works/doccano}} for labelling, which provides better support for labelling long chunks in big documents compared to other popular sequence labelling annotation tools.

We obtain labelling for 1,317 questions, totalling to 11,580 spans (including spans labelled as O) and over 1.18 million words. We divide the data into 80:10:10 train, val, and test splits, at random. High-level statistics for the dataset are presented in Table~\ref{dataset}. Figure~\ref{fig:pie} shows the average number of words per tag in the dataset. The tags \textit{Command Output} and \textit{Error Message} are relatively infrequent (1.2 and 0.6 per question) compared to the tag \textit{Command Code} (2.1 per question), however, they cover a much larger fraction of words because they tend to be quite verbose.
% Table~\ref{spanlengths} presents the average length of a span for each tag.

In Figure~\ref{fig:confusion_annotator} we show the inter-annotator agreement between two annotators on 50 questions. Few of the label pairs with large off-diagonal values include - 
\begin{itemize}
    \item \textit{Command Output} - \textit{Error Message}, which is understandable, as error messages are often interspersed in successful program runs. Conversely, unsuccessful program runs often contain a long train of success messages, only ending in one or few error logs.
    \item \textit{Command Output} - \textit{Semi-Structured Information} and \textit{File Content} - \textit{Semi-Structured Information}. This kind of confusion is due to the presence of network configurations, commands to view these, and files that contain these. They're often stored in configuration files as ``key-value" pairs
    \item \textit{Command Output} - \textit{File Content}. This particular confusion stems from the ``cat" command, and its use to view the contents of files.
\end{itemize}
The low inter-annotator agreement ($\kappa=0.7637$) illustrates the inherent difficulty of the task. At this point, it's important to note that while there's some confusion in identifying labels for these segments, the need for these separate labels stems from downstream tasks.
% \todoag{Pie chart of the number of words per tag in the dataset}
% \todoag{Table of average length of each tag}
%\todoag{Confusion matrix}

\section{Model}
% \todokc{Talk about problem formulation, modelling decisions, introduction of multiple LMs, and a subsection on the sentence level baseline}

\begin{figure}[t]
    \centering
    \includegraphics[width=0.45\textwidth]{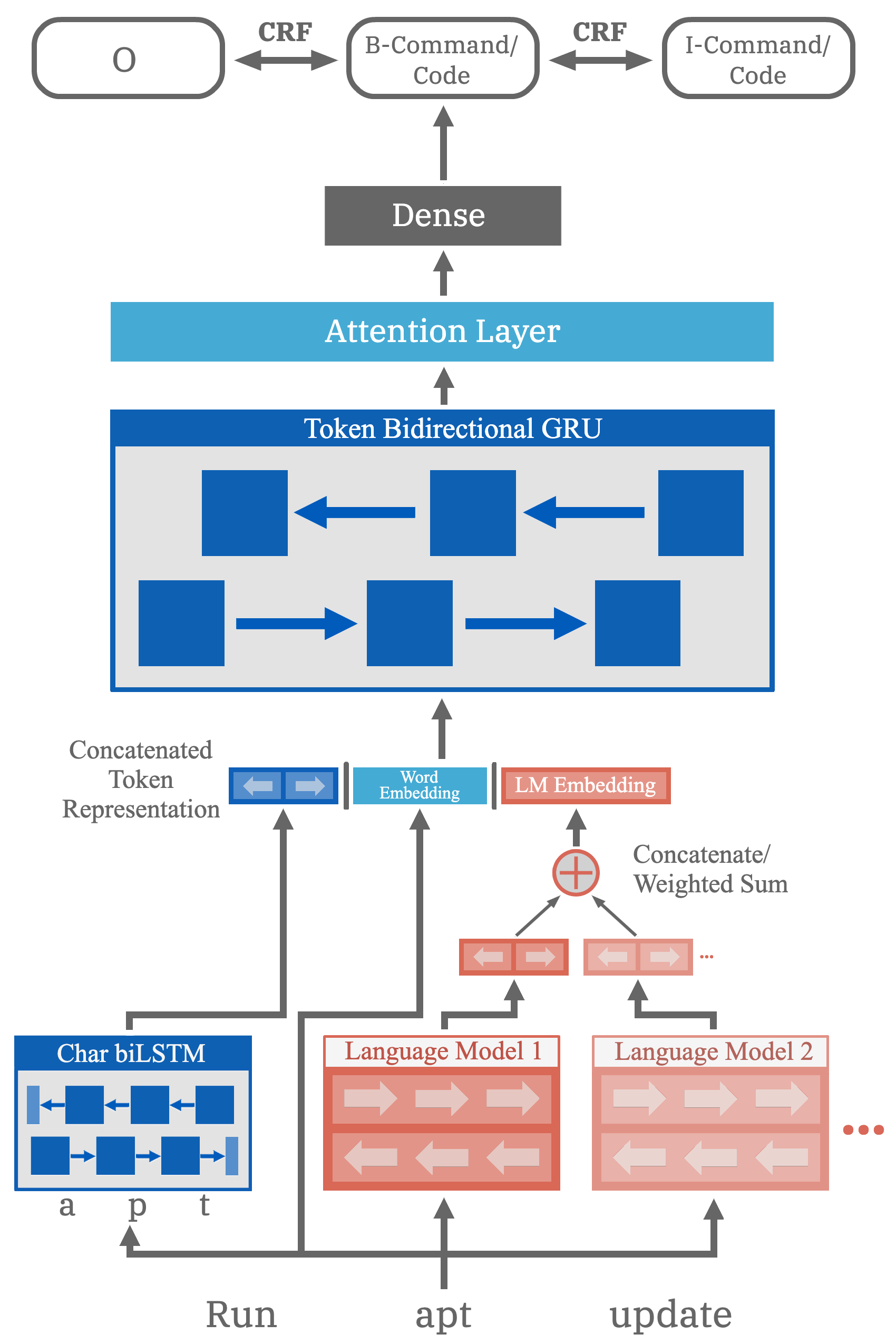}
    \caption{Model architecture for segmenting technical support problems.}
    \label{fig:Model}
\end{figure}

Given a technical support question, we formulate the segmentation problem as a sequence labelling task. It is an intuitive choice, given its efficacy for similar text segmentation problems like discourse segmentation~\cite{Wang2018} and chunking~\cite{Peters2017}. 
%The model we use for sequence labelling is most similar to ~\citet{Wang2018}.
Figure~\ref{fig:Model} presents an overview of our model. We explore different embeddings for each word (character-based embeddings, pre-trained embeddings, and pre-trained contextual embeddings). These word embeddings are then fed to a bi-directional GRU for encoding context. On the output of the GRU layer, we explore the effect of attention. Finally, the representations are passed to a CRF to decode the segment labels. 
%Since the vocabulary and grammar are distinct for each segment, we also study the impact of combining pre-trained contextual embeddings from multiple language models, trained on different data sources. In the rest of this section we detail individual components of the model.
We also study the impact of combining pre-trained contextual embeddings from multiple language models, trained on different data sources. In the rest of this section we detail individual components of the model.
% \subsection{Baseline}
% As most of the segments span across multiple lines/sentences, we choose a sentence classification model as our baseline. We use sentence embeddings obtained from the ELMo pre-trained language model as features for a 2 layer fully-connected network for classification.

% \subsection{Proposed Approach}
% We adopt biGRU-CRF based approaches and experiment with various types of distributed word representations and embeddings from language models as our input features. The key components in our model is as follows:

\subsection{Word Embeddings}
For distributed word representations, we use skip-gram based word2vec embeddings \citep{Mikolov2013} trained on all the questions from Ask Ubuntu. 
%There are multiple problems with this on our dataset, given the large volume of non-natural language text. First, due to imperfect tokenization, we frequently encounter Out Of Vocabulary tokens. Second, since a large number of tokens appear exactly once, their word vectors are probably not very accurate. 
%Given these shortcomings, we also look at fastText word embeddings \citep{Bojanowski2017}, which enrich word vectors by using subword information, emitting plausible word representations for unseen or rare words. We use a 3300-dimensional embedding from both word2vec and fastText.
We also look at fastText word embeddings \citep{Bojanowski2017}, which enrich word vectors by using subword information, emitting plausible word representations for unseen or rare words, giving us a significant gain. We use a 300-dimensional embedding from both word2vec and fastText.

\subsection{Character Embeddings}
In addition to the word-level features we also use bi-directional LSTM based character-level features similar to \citet{Chiu2016}, \citet{Lample2016}, and \citet{Ma2016}. These features encode rich character level information which can improve performance, especially in syntactic tasks. We obtain an 80-dimensional representation for each word through the character bi-LSTM, which is the concatenation of the last hidden state of the forward and backward LSTMs.

\subsection{Contextual Embeddings from Language Models}
\label{subsec:lm}
Pre-trained contextual embeddings have been shown to work well on a wide variety of NLP tasks. In domains with relatively small task-specific training data, the gains have been substantial \citep{McCann2017, Akbik2018, Peters2017}. We also include contextual embeddings from the pre-trained bi-directional language model in ELMo \citep{DBLP:journals/corr/abs-1802-05365}.

We observe that the non-natural language segments exhibit wide differences in syntactic and semantic structure, as is evident from Fig~\ref{fig:SegmentLabels}. We propose contextual embeddings from multiple language models; each trained on a different data source - English text, code snippets, config/log file contents. We hypothesize that combined embeddings from language models trained on separate data sources can capture word relationships better and can give richer word representations, as opposed to a single model trained on a large English corpora.

For combining multiple contextual embeddings, we explore two techniques - (1) a naive concatenation, and (2) a weighted sum, with weights learned from context-independent DME (Dynamic Meta-Embeddings) and context-dependent CDME (Contextualised Dynamic Meta-Embeddings) self-attention mechanisms as proposed by \citet{Kiela2018}.

\subsubsection{DME and CDME}

When using embeddings from $n$ different LMs for a training instance with $s$ tokens $\left\{\mathbf{t}_{j}\right\}_{j=1}^{s}$, we get contextual embeddings $\left\{\mathbf{w}_{i, j}\right\}_{j=1}^{s} \in \mathbb{R}^{d_{i}}(i=1,2, \ldots, n)$.

For computing the weighted sum, the embeddings from multiple LMs are first projected to a common $d'$-dimensional space by learned linear functions:
\begin{equation}
    \mathbf{w}_{i, j}^{\prime}=\mathbf{P}_{i} \mathbf{w}_{i, j}+\mathbf{b}_{i}(i=1,2, \ldots, n)
\end{equation}
where $\mathbf{P}_{i} \in \mathbb{R}^{d^{\prime} \times d_{i}}$ and $\mathbf{b}_{i} \in \mathbb{R}^{d^{\prime}}$. The projected embeddings are then combined with a weighted sum
\begin{equation}
    \mathbf{w}_{j}^{D M E}=\sum_{i=1}^{n} \alpha_{i, j} \mathbf{w}_{i, j}^{\prime}
\end{equation}
where $\alpha_{i, j}=g(\{\mathbf{w}_{i, j}^{\prime}\}_{j=1}^{s})$ are scalar weights. In DME, they are learned with the self-attention mechanism:
\begin{equation}
    \alpha_{i, j}=g\left(\mathbf{w}_{i, j}^{\prime}\right)=\phi\left(\mathbf{a} \cdot \mathbf{w}_{i, j}^{\prime}+b\right)
\end{equation}
where $\mathbf{a} \in \mathbb{R}^{d^{\prime}}$ and $b \in \mathbb{R}$ are learned parameters and $\phi$ is the softmax function.

For CDME, the self-attention mechanism is made context-dependent:
\begin{equation}
    \left.\alpha_{i, j}=g\left(\{\mathbf{w}_{i, j}^{\prime}\right\}_{j=1}^{s} \right)=\phi\left(\mathbf{a} \cdot \mathbf{h}_{j}+b\right)
\end{equation}
where $\mathbf{h}_{j} \in \mathbb{R}^{2 m}$ is the $j^{th}$ hidden state of a bi-directional LSTM which takes $\{\mathbf{w}_{i, j}^{\prime}\}_{j=1}^{s}$ as input, $\mathbf{a} \in \mathbb{R}^{2 m}$ and $b \in \mathbb{R}$. $m$ is the number of hidden units in this LSTM, and it is set to 2 as in the original paper.

\subsubsection{Data Sources for pre-trained LMs}

In addition to the pre-trained ELMo model, we train three additional language models on different data sources. Each of these are also trained with the ELMo architecture. The pre-trained model emits word embeddings of size 1024, while each of our domain-specific models emit embeddings of size 256.

\begin{itemize}
    \item \textbf{Code LM}: This LM was trained on a concatenation of all text inside the $\langle code \rangle$ tags of Ask Ubuntu, Super User, and Unix Stack Exchange posts. The total size of this corpus was approximately 400 MB.
    \item \textbf{Prog LM}: This LM was trained on a corpus containing programming source code that was compiled from various code repositories on GitHub. Approximately 275 MB in size, it includes sources in most popular languages such as C, C++, Python, Java, Go, JavaScript, and Bash. 
    \item \textbf{Config LM}: This LM was trained on a corpus of configuration and log files present in the system folders of Mac OS and Ubuntu installations. The total size of the corpus was about 60 MB.
\end{itemize}

\subsection{Attention}
In ~\citet{Wang2018}, the authors experiment with a restricted attention mechanism on top of the LSTM hidden representations. This is not appropriate for our task since the questions are fairly long (averaging around 900 words) and signals indicating the start or end of a segment might appear far away. Since RNNs are known to be poor at modelling very long-distance dependencies, we also experiment with the inclusion of the Scaled Dot-Product Attention layer \citep{Vaswani2017} on top of the bi-directional GRU. This attention layer requires the computation of 3 matrices (Key, Query, Value) from the RNN hidden states, which entails a large number of extra parameters to be learned. Therefore, we also try a version of attention where all the three matrices are set equal to the hidden states of the GRU. We call these two approaches ``weighted" and ``un-weighted" attention, in our experiments. 

\section{Experimental Setup}
With the setup above, we study the performance of various model components on the task of segmenting support problems. To put the performance in perspective, we also compare against three baselines detailed in Section~\ref{sec:baseline}. The evaluation metrics are carefully selected, avoiding an exact evaluation of such long and noisy segments, and rewarding partial retrieval of segments. The chosen evaluation metric is discussed in Section~\ref{sec:eval}. Finally, to demonstrate the usefulness of the task, we evaluate the performance of answer retrieval with segmentation (Section~\ref{sec:retrieval}).

All baselines and sequence labelling models are trained on the train split, and fine-tuned on the validation split. For the baselines, we only tune the regularization strength parameter. For the sequence labelling model, we tune the dropout and recurrent dropout parameters, as well as the learning rate. Our best performing models have a dropout of 0.3, recurrent dropout of 0, and learning rate of 1e-3. All results are then reported on the test split.

\subsection{Baseline}
\label{sec:baseline}
The task of segmenting technical support problems can be thought to be comprised of two distinct sub-tasks - (1) segmentation of text, (2) identification of the segment label. With these in mind, we propose 3 baseline methods -
%Our baseline model assumes a trivial solution for (1), by breaking text into newlines and sentences. This comes from the observation that users add newlines when presenting command outputs, error codes, or configuration files in questions. The baseline therefore focuses on solving (2) - identification of segment labels. In particular, we benchmark the performance of the following two simple baselines on the dataset -
\begin{enumerate}
    \item Sentence Only Baseline - Segmentation is done trivially with newlines and sentence boundaries serving as segment boundaries. The label for a segment is determined using just the current sentence as input.
    \item Sentence Context Baseline - Segmentation is done identically to the Sentence Only baseline. The label for a segment is determined using the immediate neighbouring sentences along with the current sentence as input.
    \item Supervised Text Segmentation Baseline - Segments are identified with the supervised algorithm for segmenting text as described in ~\citet{DBLP:journals/corr/abs-1803-09337}. The label for each segment is identified with all the text contained in it as input.
\end{enumerate}

For training the supervised text segmentation model from ~\citet{DBLP:journals/corr/abs-1803-09337} we use the whole data dump from AskUbuntu, with the $\langle code \rangle$ and $\langle / code \rangle$ html tags serving as segment boundaries.

For identifying segments (in all three baselines) we use a Logistic Regression classifier with representation from ELMo as input features. Segment representations are created by mean pooling the contextual representation of the comprising words from ELMo. 
%It's important to note that while the training and prediction on ticket happens at the sentence level, evaluation is still done per-span, to keep the results comparable with the sequence labelling model.

\subsection{Evaluation Metrics}
\label{sec:eval}
Segments in our dataset are typically quite long, therefore evaluation based on an exact match is quite harsh. Keeping this in mind, we resort to soft precision and recall metrics. We adopt proportional overlap based metrics, used for the task of opinion expression detection, as proposed by \citet{Johansson2010}. 

Towards the calculation of soft precision and recall, consider two spans $s$ and $s^{\prime}$ with labels $l$ and $l^{\prime}$ respectively. The \textit{span coverage}, $c$, is defined as how well $s'$ is covered by $s$:
\begin{equation}
    c\left(s, s^{\prime}\right) =
        \frac{\left|s \cap s^{\prime}\right|}{\left|s^{\prime}\right|} \; \text{if $l = l^{\prime}$, 0 otherwise}
\end{equation}
%In this equation, the operator $|\cdot|$ counts the no. of tokens and $\cap$ gives the set of tokens that the two spans $s$ and $s'$ have in common.
%The \textit{span coverage} is non-zero only when both spans have identical labels. 
%To penalize the model when the segment label is incorrectly predicted, we set $c\left(s, s^{\prime}\right)$ to zero when the labels corresponding the the segment spans $s$ ans $s'$ are different.

Using span coverage, the \textit{span set coverage} of a set of spans $\boldsymbol{S}$ with respect to another set of spans $\boldsymbol{S'}$ is computed as follows:
\begin{equation}
    C\left(\boldsymbol{S}, \boldsymbol{S^{\prime}}\right)=\sum_{s_{j} \in \boldsymbol{S}} \sum_{s_{k}^{\prime} \in \boldsymbol{S^{\prime}}} c\left(s_{j}, s_{k}^{\prime}\right)
\end{equation}
Using the span set coverage, we can now define the soft precision $P$ and recall $R$ of a predicted set of spans $\hat{\boldsymbol{S}}$ with respect to the gold standard set of spans $\boldsymbol{S}$:
\begin{equation}
    P(\boldsymbol{S}, \hat{\boldsymbol{S}})=\frac{C(\boldsymbol{S}, \hat{\boldsymbol{S}})}{|\hat{\boldsymbol{S}}|} \quad R(\boldsymbol{S}, \hat{\boldsymbol{S}})=\frac{C(\hat{\boldsymbol{S}}, \boldsymbol{S})}{|\boldsymbol{S}|}
\end{equation}

In this equation, the operator $|\cdot|$ counts the no. of spans in the span set. 

\subsection{Retrieval}
\label{sec:retrieval}
An important task in the automation of technical support is the retrieval of the most relevant answer document for a given ticket (from a corpus of product documentation, FAQ docs, frequent procedures). In this experiment we demonstrate the usefulness of segmenting support tickets towards this goal. We index the text of about 250,000 answers from AskUbuntu with ElasticSearch~\footnote{\href{https://www.elastic.co/products/elasticsearch}{https://www.elastic.co/products/elasticsearch}}. Answers with a large number of downvotes, and very short answers are ignored. We use questions from our annotated dataset as search queries. We then compare the retrieval performance of querying with the whole question against a query with separate fields corresponding to each segment. In the fielded query, we set different boost values for the identified segments. Boosting a specific segment of the question with a higher value causes it to have more significance in the relevance score calculation in ElasticSearch. To decide the boost values, we calculate the average percentage word overlap between a segment in the question and its correct answer from AskUbuntu on the \textit{train} and \textit{val} sets. %We average the percentage overlap among all occurrences of a segment label in a question, across all questions in the \textit{train} and \textit{val} sets.
%The correct answer for a question is then marked ``accepted answer".
To compare retrieval performance, we evaluate the Mean Reciprocal Rank (MRR) of the correct answer for questions in the \textit{test} set.
%We use Mean Reciprocal Rank as the evaluation metric. 
%Weights for each query field in the segmented query is decided by the probability of a token in each segment appearing in the correct answer (as calculated on the validation set).

\section{Results}

\begin{table*}[t]
\centering
\begin{tabular}{|p{0.5\textwidth}|l|l|l|}
\hline
     \textbf{Model} & \textbf{P} & \textbf{R}  & \textbf{F1} \\
\hline
    Sent. Only Baseline & 47.77 & 31.75 & 38.15 \\
    Sent. Context Baseline & 52.52 & 34.03 & 41.3 \\
    Supervised Text Segmentation Baseline & 44.13 & 40.43 & 42.20 \\
    & & & \\
    SL w/o LM embeddings & 74.57 & 75.51 & 75.04 \\
    SL + pre-trained ELMo & 76.88 & 74.49 & 75.67 \\
    SL + CDME combined pre-trained Embeddings & 78.30 & 79.29 & \textbf{78.80} \\
\hline
\end{tabular}
\caption{Results comparing the three baselines against variants of our sequence labelling model. The best performing variant uses CDME to combine pre-trained embeddings from multiple language models trained on different datasources.}
\label{tab:results1}
\end{table*}

Table~\ref{tab:results1} presents evaluation metrics for the three baselines against three variants of our sequence labelling model. The first variant does not use pre-trained embeddings from language models, the second uses just pre-trained ELMo, while the third combines pre-trained embeddings from multiple language models using CDME. All three variants use fastText for word embeddings (refer Section~\ref{sec:results_fastText}), character-based embeddings, and do not have attention mechanism before the final CRF layer (refer Section~\ref{sec:results_attention}).

As one would expect, the Context Baseline performs much better than the Sentence Only Baseline. The sequence labelling models, however, outperform both the baselines by a huge margin, demonstrating the effectiveness of the model on the task. 
%The sequence labelling model with pre-trained ELMo achieves a slight improvement. 
Specifically, the best performance is achieved by combining pre-trained embeddings from multiple language models trained on different data sources. It significantly outperforms the model using embeddings from a single pre-trained model on English (explored in Section~\ref{sec:results_pre-trained}).

In the following section we present results from the various model components we explored.

\subsection{Effect of fastText}
\label{sec:results_fastText}

Row 1 and 4 in Table~\ref{tab:results-fastText-attention} presents the comparison between models using word embeddings from word2vec and fastText. Both word2vec and fastText embeddings are trained on all posts in the Ask Ubuntu dataset. As we can see, fastText gives a marked improvement over using embeddings from word2vec. This is probably due to the nature of the vocabulary in our task. Since large portions of questions are spans of \textit{command output} or \textit{error messages} a lot of tokens appear very rarely. In fact, out of the 62,501 unique tokens in the dataset, 57\% appear just once, and 78\% appear 3 or fewer times. However, the characters in these tokens are probably very informative (for example ``http" in a token would signal that the token is a URL). Therefore, fastText, which uses n-grams from a token to compute embeddings, would emit more meaningful representations.

As a simple experiment, we check the similarity of two URLs from the dataset that appear just once - \textit{http://paste.ubuntu.com/1403448/} and \textit{http://paste.ubuntu.com/14545476/}. While the cosine similarity of Word2Vec vectors for the two is $-0.07$, the similarity between the fastText vectors is $0.99$.

\begin{table}[t]
\centering
\begin{tabular}{|p{0.21\textwidth}|l|l|l|}
\hline
     \textbf{Model} & \textbf{P} & \textbf{R}  & \textbf{F1} \\
\hline
    Word2Vec (w/o Attn) & 65.20 & 58.59 & 61.72 \\
    \hspace{3pt} + weighted Attn. & 62.34 & 57.0 & 59.55 \\
    \hspace{3pt} + un-weighted Attn. & 69.21 & 56.15 & 62.0 \\
    \hline
    fastText & 74.57 & 75.51 & 75.04 \\
\hline
\end{tabular}
\caption{Results for experiments between using Word2Vec and fastText embeddings. Also includes results of using attention on top of the model with Word2Vec. Since attention results were not promising, we did not repeat them with fastText.}
\label{tab:results-fastText-attention}
\end{table}

\subsection{Effect of Attention}
\label{sec:results_attention}
Given the long tickets in our dataset, and un-reasonably long lengths of spans for labels like \textit{command output} or \textit{error messages}, we explored the usefulness of attention in our model. We used the Scaled Dot-Product Attention as in ~\cite{Vaswani2017}. Rows 2 and 3 in Table~\ref{tab:results-fastText-attention} present the results of using attention. We find that weighted attention actually hurts performance. This could be because of the large number of extra parameters introduced in the calculation of Key, Value, and Query matrices. While the un-weighted version gets around this by using the bi-directional GRU hidden states as all 3 matrices, it doesn't improve results significantly either.

\subsection{Effect of Contextual Pre-Trained Embeddings}
\label{sec:results_pre-trained}
As detailed in Section~\ref{subsec:lm}, we explore the impact of pre-trained contextual embeddings. We also test our hypothesis, that combining pre-trained embeddings from different data sources would perform better on our task than using embeddings from a language model trained on a single data source. The combination is also performed in two ways - naive concatenation of embeddings from all language models, and weighted combination using DME and CDME as in~\citet{Kiela2018}. 

\begin{table}[t]
\centering
\begin{tabular}{|p{0.21\textwidth}|l|l|l|}
\hline
     \textbf{Model} & \textbf{P} & \textbf{R}  & \textbf{F1} \\
\hline
    No Pretraining & 74.57 & 75.51 & 75.04 \\
    \hline
    Simple Concat - 1 (en) & 76.88 & 74.49 & 75.67 \\
    Simple Concat - 2 (en + config) & 77.67 & 76.12 & 76.89 \\
    Simple Concat - 3 (en + code + config) & 79.64 & 77.72 & 78.67 \\
    Simple Concat - 4 (ALL) & 76.05 & 76.65 & 76.35 \\
    \hline
    DME & 77.42 & 75.82 & 76.61\\
    CDME & 78.30 & 79.29 & \textbf{78.80} \\
\hline
\end{tabular}
\caption{Results comparing the models using various pre-trained embeddings. The \textit{en} data source is the downloaded pre-trained ELMo model. For simple concatenation, we present the results for the best model at each n combinations of data sources. For example, when concatenating any 2 datasources, the \textit{en + config} combination gives the best performance.}
\label{tab:results-pretrained}
\end{table}

Table~\ref{tab:results-pretrained} summarizes these results. For the simple concatenation method, we present results for the best $n$-way combination of embeddings from different data sources, for each $n$ (1, 2, 3, and 4). We find that combining embeddings from multiple language models trained on different data sources considerably outperforms using embeddings from a single pre-trained model (using both the naive concatenation and CDME). This is an artifact of the support problems containing large sections of non-natural language text. We also find that contextual weighting does better than a simple concatenation.

\subsection{Retrieval of the Correct Answer}
\begin{table}[t]
\centering
\begin{tabular}{|p{0.35\textwidth}|l|}
\hline
     \textbf{Method} & \textbf{MRR} \\
\hline
    Full Question &  0.292\\ 
    Segmented Question - Gold &  0.300\\ 
    Segmented Question - Predicted &  \textbf{0.298}\\ 
\hline
\end{tabular}
\caption{Retrieval results, comparing the performance of querying with the full question against segmented question (gold segments and predicted segments)}
\label{tab:results-retrieval}
\end{table}
\label{sec:results_retrieval}
Table~\ref{tab:results-retrieval} presents results for the retrieval experiment. We show that weighing identified segments of the question with separate weights improves retrieval of the correct answer over a query with all tokens from the question. We also present results from the gold annotations of segments for these questions, as an upper-bound of the performance improvement we can hope to achieve.

% Performance against sentence baseline

% 0. Effect of multiple LMs

% 1. Confusion and mistakes from the model

% 2. Long-distance dependencies? - Experiments to show if we're capturing, or not capturing these as expected

% 3. Attention? - One way to handle long-distance dependencies -> not effective

% 4. Ablation - What does a model without the various parts do? To understand what these parts contribute -

% a. CRF

% b. Word vectors

% c. Character Embeddings

\section{Conclusion}

% Customer support for technical products is a huge industry, with high costs involved in hiring and training staff. Automation, to assist human agents with easily solvable problems, would reduce this strain. Understanding technical support tickets is a crucial part of that effort. 
In this paper, we introduce and address an important problem towards a better understanding of support tickets - segmentation of various non-natural language segments. We create an annotated dataset for the task, on questions from the publicly available website, Ask Ubuntu. We also study the performance of the most recent Recurrent Neural Network-based approaches to sequence labelling, on this task. In the end, we propose the novel idea of combining pre-trained embeddings from language models trained on different data sources, which substantially improves performance. We also demonstrate the usefulness of the task with improvements in retrieval of the correct answer. Our future research direction includes a thorough study of differences in this dataset with actual tickets, and potential for transfer. It is still valuable to study models on open datasets, however, as these are readily available to the community.

\bibliography{ticketseg}

\begin{thebibliography}{22}
\expandafter\ifx\csname natexlab\endcsname\relax\def\natexlab#1{#1}\fi

\bibitem[{Akbik et~al.(2018)Akbik, Blythe, and Vollgraf}]{Akbik2018}
Alan Akbik, Duncan Blythe, and Roland Vollgraf. 2018.
\newblock \href {https://www.aclweb.org/anthology/C18-1139} {Contextual string
  embeddings for sequence labeling}.
\newblock In \emph{Proceedings of the 27th International Conference on
  Computational Linguistics}, pages 1638--1649, Santa Fe, New Mexico, USA.
  Association for Computational Linguistics.

\bibitem[{Bojanowski et~al.(2017)Bojanowski, Grave, Joulin, and
  Mikolov}]{Bojanowski2017}
Piotr Bojanowski, Edouard Grave, Armand Joulin, and Tomas Mikolov. 2017.
\newblock \href {https://doi.org/10.1162/tacl_a_00051} {Enriching word vectors
  with subword information}.
\newblock \emph{Transactions of the Association for Computational Linguistics},
  5:135--146.

\bibitem[{Chiu and Nichols(2016)}]{Chiu2016}
Jason~P.C. Chiu and Eric Nichols. 2016.
\newblock \href {https://doi.org/10.1162/tacl_a_00104} {Named entity
  recognition with bidirectional {LSTM}-{CNN}s}.
\newblock \emph{Transactions of the Association for Computational Linguistics},
  4:357--370.

\bibitem[{Choi(2000)}]{choi-2000-advances}
Freddy Y.~Y. Choi. 2000.
\newblock \href {https://www.aclweb.org/anthology/A00-2004} {Advances in domain
  independent linear text segmentation}.
\newblock In \emph{Proceedings of the 1st North American Chapter of the
  Association for Computational Linguistics Conference}, NAACL 2000, page
  26–33, USA. Association for Computational Linguistics.

\bibitem[{{Chrupa{\l}a}(2013)}]{Chrupala2013}
Grzegorz {Chrupa{\l}a}. 2013.
\newblock \href {http://arxiv.org/abs/1309.4628} {Text segmentation with
  character-level text embeddings}.
\newblock Workshop on Deep Learning for Audio, Speech and Language Processing,
  ICML 2013, Atlanta, United States.

\bibitem[{Gupta et~al.(2018)Gupta, Ray, Dasgupta, Singh, Aggarwal, and
  Mohapatra}]{gupta-etal-2018-semantic}
Abhirut Gupta, Anupama Ray, Gargi Dasgupta, Gautam Singh, Pooja Aggarwal, and
  Prateeti Mohapatra. 2018.
\newblock \href {https://www.aclweb.org/anthology/C18-1275} {Semantic parsing
  for technical support questions}.
\newblock In \emph{Proceedings of the 27th International Conference on
  Computational Linguistics}, pages 3251--3259, Santa Fe, New Mexico, USA.
  Association for Computational Linguistics.

\bibitem[{He et~al.(2016)He, Zhu, He, and Lyu}]{he2016experience}
Shilin He, Jieming Zhu, Pinjia He, and Michael~R Lyu. 2016.
\newblock \href {https://ieeexplore.ieee.org/document/7774521} {Experience
  report: system log analysis for anomaly detection}.
\newblock In \emph{2016 IEEE 27th International Symposium on Software
  Reliability Engineering (ISSRE)}, pages 207--218. IEEE.

\bibitem[{Hearst(1997)}]{Hearst1997TextTilingST}
Marti~A. Hearst. 1997.
\newblock \href {https://www.aclweb.org/anthology/J97-1003/} {Text{T}iling:
  Segmenting text into multi-paragraph subtopic passages}.
\newblock \emph{Computational Linguistics}, 23(1):33–64.

\bibitem[{{Huang} et~al.(2015){Huang}, {Xu}, and {Yu}}]{Huang2015}
Zhiheng {Huang}, Wei {Xu}, and Kai {Yu}. 2015.
\newblock \href {http://arxiv.org/abs/1508.01991} {{Bidirectional LSTM-CRF
  Models for Sequence Tagging}}.
\newblock \emph{arXiv e-prints}, page arXiv:1508.01991.

\bibitem[{Johansson and Moschitti(2010)}]{Johansson2010}
Richard Johansson and Alessandro Moschitti. 2010.
\newblock \href {https://www.aclweb.org/anthology/W10-2910} {Syntactic and
  semantic structure for opinion expression detection}.
\newblock In \emph{Proceedings of the Fourteenth Conference on Computational
  Natural Language Learning}, CoNLL ’10, page 67–76, USA. Association for
  Computational Linguistics.

\bibitem[{Kiela et~al.(2018)Kiela, Wang, and Cho}]{Kiela2018}
Douwe Kiela, Changhan Wang, and Kyunghyun Cho. 2018.
\newblock \href {https://doi.org/10.18653/v1/D18-1176} {Dynamic meta-embeddings
  for improved sentence representations}.
\newblock In \emph{Proceedings of the 2018 Conference on Empirical Methods in
  Natural Language Processing}, pages 1466--1477, Brussels, Belgium.
  Association for Computational Linguistics.

\bibitem[{Koshorek et~al.(2018)Koshorek, Cohen, Mor, Rotman, and
  Berant}]{DBLP:journals/corr/abs-1803-09337}
Omri Koshorek, Adir Cohen, Noam Mor, Michael Rotman, and Jonathan Berant. 2018.
\newblock \href {https://doi.org/10.18653/v1/N18-2075} {Text segmentation as a
  supervised learning task}.
\newblock In \emph{Proceedings of the 2018 Conference of the North {A}merican
  Chapter of the Association for Computational Linguistics: Human Language
  Technologies, Volume 2 (Short Papers)}, pages 469--473, New Orleans,
  Louisiana. Association for Computational Linguistics.

\bibitem[{Lample et~al.(2016)Lample, Ballesteros, Subramanian, Kawakami, and
  Dyer}]{Lample2016}
Guillaume Lample, Miguel Ballesteros, Sandeep Subramanian, Kazuya Kawakami, and
  Chris Dyer. 2016.
\newblock \href {https://doi.org/10.18653/v1/N16-1030} {Neural architectures
  for named entity recognition}.
\newblock In \emph{Proceedings of the 2016 Conference of the North {A}merican
  Chapter of the Association for Computational Linguistics: Human Language
  Technologies}, pages 260--270, San Diego, California. Association for
  Computational Linguistics.

\bibitem[{Ma and Hovy(2016)}]{Ma2016}
Xuezhe Ma and Eduard Hovy. 2016.
\newblock \href {https://doi.org/10.18653/v1/P16-1101} {End-to-end sequence
  labeling via bi-directional {LSTM}-{CNN}s-{CRF}}.
\newblock In \emph{Proceedings of the 54th Annual Meeting of the Association
  for Computational Linguistics (Volume 1: Long Papers)}, pages 1064--1074,
  Berlin, Germany. Association for Computational Linguistics.

\bibitem[{McCann et~al.(2017)McCann, Bradbury, Xiong, and Socher}]{McCann2017}
Bryan McCann, James Bradbury, Caiming Xiong, and Richard Socher. 2017.
\newblock \href {http://dl.acm.org/citation.cfm?id=3295222.3295377} {Learned in
  translation: Contextualized word vectors}.
\newblock In \emph{Proceedings of the 31st International Conference on Neural
  Information Processing Systems}, NIPS’17, page 6297–6308, Red Hook, NY,
  USA. Curran Associates Inc.

\bibitem[{Mikolov et~al.(2013)Mikolov, Sutskever, Chen, Corrado, and
  Dean}]{Mikolov2013}
Tomas Mikolov, Ilya Sutskever, Kai Chen, Greg Corrado, and Jeffrey Dean. 2013.
\newblock \href {http://dl.acm.org/citation.cfm?id=2999792.2999959}
  {Distributed representations of words and phrases and their
  compositionality}.
\newblock In \emph{Proceedings of the 26th International Conference on Neural
  Information Processing Systems - Volume 2}, NIPS'13, pages 3111--3119, USA.
  Curran Associates Inc.

\bibitem[{Peters et~al.(2017)Peters, Ammar, Bhagavatula, and
  Power}]{Peters2017}
Matthew Peters, Waleed Ammar, Chandra Bhagavatula, and Russell Power. 2017.
\newblock \href {https://doi.org/10.18653/v1/P17-1161} {Semi-supervised
  sequence tagging with bidirectional language models}.
\newblock In \emph{Proceedings of the 55th Annual Meeting of the Association
  for Computational Linguistics (Volume 1: Long Papers)}, pages 1756--1765,
  Vancouver, Canada. Association for Computational Linguistics.

\bibitem[{Peters et~al.(2018)Peters, Neumann, Iyyer, Gardner, Clark, Lee, and
  Zettlemoyer}]{DBLP:journals/corr/abs-1802-05365}
Matthew Peters, Mark Neumann, Mohit Iyyer, Matt Gardner, Christopher Clark,
  Kenton Lee, and Luke Zettlemoyer. 2018.
\newblock \href {https://doi.org/10.18653/v1/N18-1202} {Deep contextualized
  word representations}.
\newblock In \emph{Proceedings of the 2018 Conference of the North {A}merican
  Chapter of the Association for Computational Linguistics: Human Language
  Technologies, Volume 1 (Long Papers)}, pages 2227--2237, New Orleans,
  Louisiana. Association for Computational Linguistics.

\bibitem[{Rei et~al.(2016)Rei, Crichton, and Pyysalo}]{Rei2016}
Marek Rei, Gamal Crichton, and Sampo Pyysalo. 2016.
\newblock \href {https://www.aclweb.org/anthology/C16-1030} {Attending to
  characters in neural sequence labeling models}.
\newblock In \emph{Proceedings of {COLING} 2016, the 26th International
  Conference on Computational Linguistics: Technical Papers}, pages 309--318,
  Osaka, Japan. The COLING 2016 Organizing Committee.

\bibitem[{Vaswani et~al.(2017)Vaswani, Shazeer, Parmar, Uszkoreit, Jones,
  Gomez, Kaiser, and Polosukhin}]{Vaswani2017}
Ashish Vaswani, Noam Shazeer, Niki Parmar, Jakob Uszkoreit, Llion Jones,
  Aidan~N Gomez, \L~ukasz Kaiser, and Illia Polosukhin. 2017.
\newblock \href
  {http://papers.nips.cc/paper/7181-attention-is-all-you-need.pdf} {Attention
  is all you need}.
\newblock In I.~Guyon, U.~V. Luxburg, S.~Bengio, H.~Wallach, R.~Fergus,
  S.~Vishwanathan, and R.~Garnett, editors, \emph{Advances in Neural
  Information Processing Systems 30}, pages 5998--6008. Curran Associates, Inc.

\bibitem[{Wang et~al.(2010)Wang, Ming, Hu, and Chua}]{Wang2010}
Kai Wang, Zhao-Yan Ming, Xia Hu, and Tat-Seng Chua. 2010.
\newblock \href {https://doi.org/10.1145/1835449.1835515} {Segmentation of
  multi-sentence questions: Towards effective question retrieval in {CQA}
  services}.
\newblock In \emph{Proceedings of the 33rd International ACM SIGIR Conference
  on Research and Development in Information Retrieval}, SIGIR ’10, page
  387–394, New York, NY, USA. Association for Computing Machinery.

\bibitem[{Wang et~al.(2018)Wang, Li, and Yang}]{Wang2018}
Yizhong Wang, Sujian Li, and Jingfeng Yang. 2018.
\newblock \href {https://www.aclweb.org/anthology/D18-1116} {Toward fast and
  accurate neural discourse segmentation}.
\newblock In \emph{Proceedings of the 2018 Conference on Empirical Methods in
  Natural Language Processing}, pages 962--967, Brussels, Belgium. Association
  for Computational Linguistics.

\end{thebibliography}
\bibliographystyle{acl_natbib}

\end{document}